\documentclass[11pt,a4paper]{article}
\usepackage[hyperref]{acl2020}
\usepackage{times}
\usepackage{latexsym}
\renewcommand{\UrlFont}{\ttfamily\small}
\usepackage{microtype}

\aclfinalcopy 

\usepackage{csquotes}    
\usepackage{multirow}
\usepackage{graphicx}    
\usepackage{booktabs}    
\usepackage{xcolor}      
\usepackage{tikz}        

\newcommand{\DEVELOPMENT}{1} 
\usepackage{ifthen}
\ifthenelse{\DEVELOPMENT = 1}{
	\newcommand{\tz}[1]{\textcolor{purple}{\textbf{TZ:} #1}}
    \newcommand{\obe}[1]{\textcolor{blue}{\textbf{OE:} #1}}
}{
	\newcommand{\tz}[1]{}
	\newcommand{\obe}[1]{}
}

\makeatletter
\def\url@leostyle{%
  \@ifundefined{selectfont}{\def\UrlFont{\sf}}{\def\UrlFont{\scriptsize\sffamily}}}
\makeatother
\MakeOuterQuote{"}
\definecolor{e1}{HTML}{DF7162}
\definecolor{e2}{HTML}{B2E684}
\definecolor{e3}{HTML}{4BC5BE}
\definecolor{e4}{HTML}{E38E48}
\definecolor{e5}{HTML}{6A9BDD}

\title{Effects of Layer Freezing on Transferring a\\Speech Recognition System to Under-resourced Languages}

\author{Onno Eberhard \textnormal{and} Torsten Zesch\\
  Language Technology Lab \\
  University of Duisburg-Essen, Germany \\
  \texttt{onno.eberhard@stud.uni-due.de} \\
  \texttt{torsten.zesch@uni-due.de}}

\date{}

\begin{document}
\maketitle
\begin{abstract}
In this paper, we investigate the effect of layer freezing on the effectiveness of model transfer in the area of automatic speech recognition.
We experiment with Mozilla's DeepSpeech architecture on German and Swiss German speech datasets and compare the results of either training from scratch vs.\ transferring a pre-trained model.
We compare different layer freezing schemes and find that even freezing only one layer already significantly improves results.
\end{abstract}

\section{Introduction}
The field of automatic speech recognition (ASR) is dominated by research specific to the English language. 
There exist plenty available text-to-speech models pre-trained on (and optimized for) English data. 
When it comes to a low-resource language like Swiss German, or even standard German, only a very limited number of small-scale models is available. 
In this paper, we train Mozilla's implementation\footnote{\url{https://github.com/mozilla/DeepSpeech}} of Baidu's DeepSpeech ASR architecture \citep{hannun2014deep} on these two languages. 
We use transfer learning to leverage the availability of a pre-trained English version of DeepSpeech and observe the difference made by freezing different numbers of layers during training.


\section{Transfer Learning and Layer Freezing}
Deep neural networks can excel at many different tasks, but they often require very large amounts of training data and computational resources.
To remedy this, it is often advantageous to employ transfer learning: Instead of initializing the parameters of the network randomly, the optimized parameters of a network trained on a similar task are reused.
Those parameters can then be fine-tuned to the specific task at hand, using less data and fewer computational resources.
In the fine-tuning process many parameters of the original model may be ``frozen'', i.e.\ held constant during training.
This can speed up training and improve results when less training data is available \citep{DBLP:conf/rep4nlp/KunzeKKKJS17}.
The idea of taking deep neural networks trained on large datasets and fine-tuning them on tasks with less available training data has been popular in computer vision for years \citep{huh2016makes}.
More recently, with the emergence of end-to-end deep neural networks for automatic speech recognition (like DeepSpeech), it has also been used in this area \citep{DBLP:conf/rep4nlp/KunzeKKKJS17, DBLP:journals/corr/abs-1911-09271}.

Deep neural networks learn representations of the input data in a hierarchical manner.
The input is transformed into simplistic features in the first layers of a neural network and into more complex features in the layers closer to the output.
If we assume the simplistic feature representations are applicable in similar, but different, contexts, layer-wise freezing of parameters seems like a good choice. This is further reinforced by findings from image classification \citep{DBLP:conf/nips/YosinskiCBL14}, where the learned features can additionally be nicely visualized \citep{10.1007/978-3-319-10590-1_53}.

As for automatic speech recognition, the representations learned by the layers is not as clear-cut as within image processing.
Nonetheless, some findings, for example that affricates are better represented at later layers in the network \citep{NIPS2017}, seem to affirm the hypothesis that the later layers learn more abstract features and earlier layers learn more primitive features. 
This is important for fine-tuning, because it only makes sense to freeze parameters if they don't need to be adjusted for the new task. 
If it is known that the first layers of a network learn to identify ``lower-level''-features, i.e.\ simple shapes in the context of image processing or simple sounds in the context of ASR, these layers can be frozen completely during fine-tuning.


\section{Experimental Setup}

In our experiments, we transfer an English pre-trained version of DeepSpeech to German and to Swiss German data and observe the impact of freezing fewer or more layers during training.

\subsection{Datasets}

\begin{table}[t]
    \small
    \centering
    \begin{tabular}{llrrr}
        \toprule
        & Dataset & Hours & Speakers\\
        \midrule
        Pre-training & English & $>$6,500 & ? \\
        \addlinespace[2mm]
        \multirow{2}{*}{Transfer} & German & 315 & 4,823 \\
        & Swiss German & 70 & 191 \\
        \bottomrule
    \end{tabular}
    \caption{Overview of datasets}
    \label{tab:datasets}
\end{table}

We trained the models for (standard) German on the German part of the Mozilla Common Voice speech dataset \citep{DBLP:conf/lrec/ArdilaBDKMHMSTW20}. 
The utterances are typically between 3 and 5 seconds long and are collected from and reviewed by volunteers.
This collection method entails a rather high number of speakers and quite some noise.
The Swiss German models were trained on the data provided by \citet{pluess}. 
This speech data was collected from speeches at the Bernese parliament. 
The English pre-trained model was trained by Mozilla on a combination of English speech datasets, including LibriSpeech and Common Voice English.\footnote{\url{https://github.com/mozilla/DeepSpeech/releases/tag/v0.7.0}} 
The datasets for all three languages are described in Table~\ref{tab:datasets}. 
For inference and testing we used the language model KenLM \citep{heafield-2011-kenlm}, trained on the corpus described by \citet[Section~3.2]{Radeck-Arneth2015}. 
This corpus consists of a mixture of texts from the sources Wikipedia and Europarl as well as crawled sentences. 
The whole corpus was preprocessed with MaryTTS \citep{schroder2003german}.

\subsection{ASR Architecture}
\begin{figure}[t]
    \centering
    \input{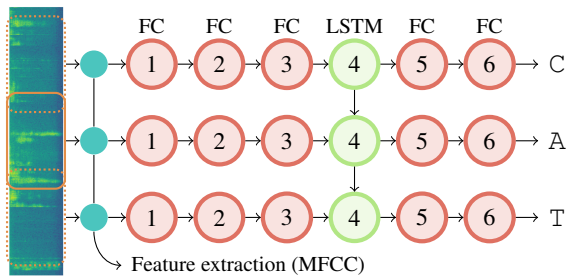}
    \caption{DeepSpeech architecture. The fully connected (FC) layers 1 -- 3 and 5 are ReLU activated, the last layer uses a softmax function to compute character probabilities.}
    \label{fig:ds}
\end{figure}

We use Mozilla's DeepSpeech version 0.7 for our experiments.
The implementation differs in many ways from the original model presented by \citet{hannun2014deep}. 
The architecture is described in detail in the official documentation\footnote{\url{https://deepspeech.readthedocs.io/en/latest/DeepSpeech.html}} and is depicted in Figure~\ref{fig:ds}.
From the raw speech data, Mel-Frequency Cepstral Coefficients \citep{imai1983cepstral} are extracted and passed to a 6-layer deep recurrent neural network. 
The first three layers are fully connected with a ReLU activation function. 
The fourth layer is a Long Short-Term Memory (LSTM) unit \citep{hochreiter1997long}; the fifth layer is again fully connected and ReLU activated. 
The last layer outputs probabilities for each character in the language's alphabet. 
It is fully connected and uses a softmax activation for normalization. 
The character-probabilities are used to calculate a Connectionist Temporal Classification (CTC) loss function \citep{graves2006connectionist}.
The weights of the model are optimized using the Adam method \citep{kingma2014adam} with respect to the CTC loss.

\subsection{Training Details} \label{sec:training}
As a baseline, we directly train the German and Swiss German model on the available data from scratch, without any transfer (hereafter called "Baseline"). To assess the effects of layer freezing, we then re-train the model based on weight initialization from the English pre-trained model.\footnote{\url{https://github.com/mozilla/DeepSpeech/releases}} In this step, we freeze the first $N$ layers during training, where $N = 0, \dots, 5$. For $N = 4$ we additionally experiment with freezing the 5\textsuperscript{th} layer instead of the LSTM layer, which we denote as "Layers 1-3,5 Frozen". We do this because we see the LSTM as the most essential and flexible part of the architecture; the 5\textsuperscript{th} and 6\textsuperscript{th} layer have a simpler interpretation as transforming the LSTM hidden state into character-level information. This stage should be equivalent across languages, as long as the LSTM hidden state is learned accordingly, which is ensured by not freezing the LSTM. For all models, we reinitialize the last layer, because of the different alphabet sizes of German / Swiss German and English (ä, ö, ü), but don't reinitialize any other layers (as done e.g. by \citet{hjortnaes-etal-2020-towards}). The complete training script, as well as the modified versions of DeepSpeech that utilize layer freezing are available online\footnotemark. The weights were frozen by adding \texttt{trainable=False} at the appropriate places in the TensorFlow code, though some other custom modifications were necessary and are described online\footnotemark[\value{footnote}]. For Swiss German, we do not train the network on the German dataset first and transfer from German to Swiss German, as this has been shown to lead to worse results \citep{agarwal2020ltl}. \footnotetext{\url{https://github.com/onnoeberhard/deepspeech}}

\subsection{Hyperparameters \& Server}
In training each model, we used a batch size of 24, a learning rate of 0.0005 and a dropout rate of 0.4. 
We did not perform any hyperparameter optimization. 
The training was done on a Linux machine with 96 Intel Xeon Platinum 8160 CPUs @ 2.10GHz, 256GB of memory and an NVIDIA GeForce GTX 1080 Ti GPU with 11GB of memory. 
Training the German language models for 30 epochs took approximately one hour per model. 
Training the Swiss German models took about 4 hours for 30 epochs on each model.
We did not observe a correlation between training time and the number of frozen layers.
For testing, the epoch with the best validation loss during training was taken for each model. 

\section{Results \& Discussion}
Results of our baselines are very close to the values reported for German by \citet{agarwal-zesch-2019-german} and Swiss German by \citet{agarwal2020ltl} using the same architecture.

\begin{figure}[t]
    \centering
    \includegraphics{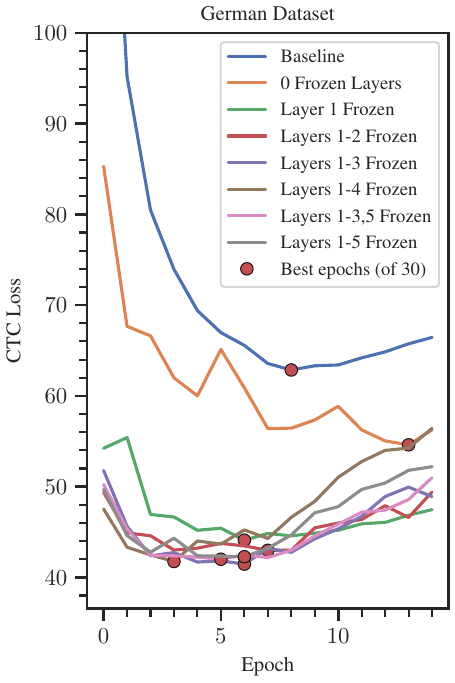}
    \caption{Learning curves (validation loss) on the German dataset. Layer freezing has a noticeable impact, but how many layers are frozen does not seem to make much of a difference. See Section \ref{sec:training} for details.}
    \label{fig:de}
\end{figure}

\begin{figure}[htb]
    \centering
    \includegraphics{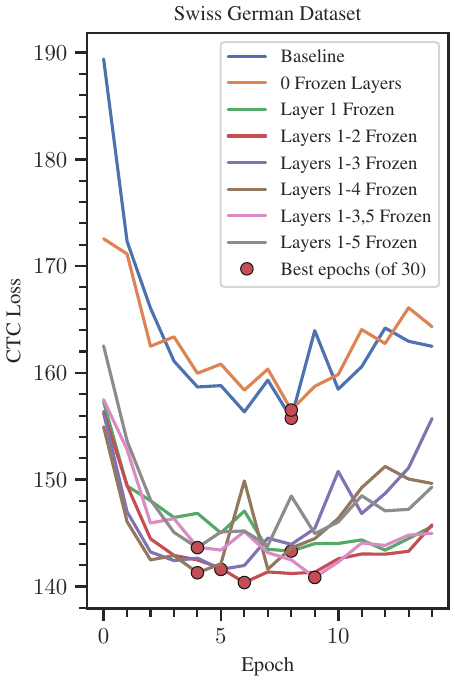}
    \caption{Learning curves (validation loss) on the Swiss German dataset. Compare with Figure \ref{fig:de}.}
    \label{fig:ch}
\end{figure}

The test results for both languages from the different models described in Section~\ref{sec:training} are compiled in Table~\ref{tab:results}.
Figures~\ref{fig:de} and \ref{fig:ch} show the learning curves for all training procedures for German and Swiss German, respectively.
The epochs used for testing (cf. Table \ref{tab:results}) are also marked in the figures.

\addtolength{\tabcolsep}{-1pt}
\begin{table}[t]
    \centering
    \begin{tabular}{lrrrr}
        \toprule
        {} & \multicolumn{2}{c}{German}  & \multicolumn{2}{c}{Swiss}\\
        \cmidrule(lr){2-3} \cmidrule(lr){4-5}
        Method & WER & CER & WER & CER\\
        \midrule
        Baseline & .70 & .42 & .74 & .52 \\
        \addlinespace[1mm]
        0 Frozen Layers & .63 & .37 & .76 & .54 \\
        Layer 1 Frozen & .48 & .26 & .69 & .48 \\
        Layers 1-2 Frozen & \bf{.44} & \bf{.22} & \bf{.67} & \bf{.45} \\
        Layers 1-3 Frozen & \bf{.44} & \bf{.22} & .68 & .47 \\
        Layers 1-4 Frozen & .45 & .24 & .68 & .47 \\
        Layers 1-3,5 Frozen & .46 & .25 & .68 & .46 \\
        Layers 1-5 Frozen & \bf{.44} & .23 & .70 & .48 \\
        \bottomrule
    \end{tabular}
    \caption{Results on test sets (cf. Section \ref{sec:training})}
    \label{tab:results}
\end{table}
\addtolength{\tabcolsep}{1pt}

For both languages, the best results were achieved by the models with the first two to three layers frozen during training. 
It is notable however, that the other models that utilize layer freezing are not far off, the learning curves look remarkably similar (in both plots, these are the lower six curves).
For both languages, these models achieve much better results than the two models without layer freezing ("Baseline" and "0 Frozen Layers").
The results seem to indicate that freezing the first layer brings the largest advantage in training, with diminishing returns on freezing the second and third layers.
For German, additionally freezing the fourth or fifth layer slightly worsens the result, though interestingly, freezing both results in better error rates. This might however only be due to statistic fluctuations, as it can be seen in Figure \ref{fig:de} that on the validation set, the model with 5 frozen layers performs worse than those with 3 or 4 frozen layers.
For Swiss German, the result slightly worsens when the third layer is frozen and performance further drops when freezing subsequent layers.
Similar results were achieved by \citet{DBLP:conf/lrec/ArdilaBDKMHMSTW20}, where freezing two or three layers also achieved the best transfer results for German, with a word error rate of 44\%.
They also used DeepSpeech and a different version of the German Common Voice dataset.


The results don't show a significant difference between freezing the fourth or the fifth layer of the network ("Layers 1-4 Frozen" vs. "Layers 1-3,5 Frozen"). This indicates that the features learned by the LSTM are not as language-specific as we hypothesized. It might even be that, in general, it does not matter much which specific layers are frozen, if the number of frozen parameters is the same. It might be interesting to see what happens if the last instead of the first layers are frozen (not necessarily with this architecture), thereby breaking the motivation of hierarchically learned features, with later layers being more task-specific.

It is interesting that the models with four or five frozen layers, i.e. only 2 or 1 learnable layers, still achieve good results. This indicates that the features extracted by DeepSpeech when trained on English are general enough to really be applicable for other languages as well.
It is probable that with a larger dataset the benefits of freezing weights decrease and better results are achieved with freezing fewer or no layers.
For both languages it is evident that the transfer learning approach is promising.

\paragraph{Limitations}
Our experiment is limited to a transfer between closely related languages.
For example, when just transcribing speech there is no need for such a model to learn intonation features.
This might be a problem when trying to transfer such a pre-trained model to a tonal language like Mandarin or Thai. 
There might also be phonemes that don't exist or are very rare in English but abundant in other languages.

\section{Summary}
We investigate the effect of layer freezing on the effectiveness of transferring a speech recognition model to a new language with limited training data.
We find that transfer is not very effective without layer freezing, but that already one frozen layer yields quite good results. The differences between freezing schemes are surprisingly small, even when freezing all layers but the last.

\section*{Acknowledgements}
We want to thank Aashish Agarwal for valuable help in setting up DeepSpeech and for providing preprocessing scripts as well as the hyperparameters we used for training.

\bibliography{anthology,acl2020,bib}
\bibliographystyle{acl_natbib}

\end{document}